\documentclass[journal]{IEEEtran}
\usepackage{cite}
\usepackage{xcolor}
\usepackage{amsmath}
\usepackage{graphicx}  
\usepackage{float}  
\usepackage{subfigure}  
\usepackage{multirow} 
\usepackage{multicol}
\usepackage{booktabs}
\usepackage{caption}
\usepackage{algorithm}
\usepackage{algpseudocode}
\usepackage{tikz}
\usepackage{flushend}
\usepackage{afterpage}
\usepackage{stfloats}
\definecolor{myred}{RGB}{192,80,70}
\definecolor{mygreen}{RGB}{157,187,97}

\begin{document}

\title{Collaborative Goal Tracking of Multiple Mobile Robots Based on Geometric Graph Neural Network}

\author{Weining Lu$^{*1}$, Qingquan~Lin$^1$, Litong Meng, Chenxi Li, Bin Liang

\thanks{* Corresponding author: Weining Lu is with Beijing National Research Center for Information Science and Technology, Tsinghua University, Beijing, China  e-mail:luwn@tsinghua.edu.cn}
\thanks{ Chenxi Li, Qingquan Lin, Litong Meng and Bin Liang are
 with the Department of Automation, Tsinghua University, Beijing,
 100084, China. (e-mail: lcx22@mails.tsinghua.edu.cn,
 linqq19@tsinghua.org.cn, menglt@ieee.org,
 liangbin@tsinghua.edu.cn).}
\thanks{Weining Lu and Qingquan Lin contribute equal to this paper.}}
\maketitle

\begin{abstract}
Multiple mobile robots play a significant role in various spatially distributed tasks.
In unfamiliar and non-repetitive scenarios, reconstructing the global map is time-inefficient and sometimes unrealistic. Hence,  research has focused on achieving real-time collaborative planning by utilizing sensor data from multiple robots located at different positions, all without relying on a global map.This paper introduces a Multi-Robot collaborative Path Planning method based on Geometric Graph Neural Network (MRPP-GeoGNN). We extract the features of each neighboring robot's sensory data and integrate the relative positions of neighboring robots into each interaction layer to incorporate obstacle information along with location details using geometric feature encoders.After that, a MLP layer is used to map the amalgamated local features to multiple forward directions for the robot's actual movement. We generated expert data in ROS to train the network and carried out both simulations and physical experiments to validate the effectiveness of the proposed method. Simulation results demonstrate an approximate $5\%$ improvement in accuracy compared to the model based solely on CNN on expert datasets. The success rate is enhanced by about $4\%$ compared to CNN, and the flowtime increase is reduced by approximately $18\%$ in the ROS test, surpassing other GNN models. Besides, the proposed method is able to leverage neighbor's information and greatly improves path efficiency in real-world scenarios.
\end{abstract}

\IEEEpeerreviewmaketitle

\section{Introduction}
\IEEEPARstart{R}{ecent} years have seen the widespread application of multi-robot systems in spatially distributed tasks including logistics\cite{multirobot_logistics}, agriculture\cite{multirobot-agriculture}, transport systems\cite{multirobot-transport} and so on.
In a multi-robot system, different robots are directly or indirectly connected to each other through the network and collaborate with each other to accomplish common or different goals that is not possible for single robot.
Path planning of multiple robots is fundamental to cooperation in task execution process , therefore the Multi-Robot Path Planning problem(MRPP) , aimed at finding obstacle-free paths to the desired end point for all robots, has attracted extensive research for a long time\cite{pathplan_review}.
MRPP is an NP-hard problem in optimization and  has high computational complexity due to multi-objective , which results in a lack of complete algorithms that offer both solution optimality and computational efficiency\cite{review_2022_machines}.

In unfamiliar and non-repetitive scenarios, reconstructing the global map is time-inefficient and sometimes unrealistic. This paper focuses on the problem of multi-robot path planning under local observation and communication conditions, with no global positional information. Compared with single robot path planning,multiple robots interconnected through the network in shared environments have the advantage of more sensing information at the same time point, which can be leveraged to generate more efficient routes when the global map is absent.
Motivated by this intuition, various methods have been proposed to tackle the MRPP problem, including artificial potential field method\cite{2019AatificialPotentialField}, bio-inspired methods like particle swarm optimization \cite{21jumpPSO}, learning based methods like multi-agent reinforcement learning\cite{2020marl} and so forth. Despite all these methods, how to utilize the high dimensional sensory information of other robots and their spatial relationships to assist path planning remains an open problem. 

In recent years, graph neural networks(GNNs) have been extensively studied due to their powerful representation capabilities for graph-structured data, which also have been applied to solve the MRPP problem. There exists different ways to utilize GNN for MRPP problems: some use GNNs to directly generate movement actions\cite{LiQb_2020GNN, LiQb_2021GAT,dhq_graphsage}, others use it to process input data and obtain an abstract representation of the current state, then combine with other methods to complete path planning tasks\cite{GnnRL_multi, 2022ICRA_gnn_policy}. However, most research is limited to ideal environments described in grid form, where the expert data gathered still exhibit discrepancies with real-world data, such as varying obstacle shapes and occlusion of LiDAR sensors.

In this work, we introduces a Multi-Robot collaborative Path Planning method
based on Geometric Graph Neural Network (MRPP-GeoGNN) for practical applications with only local observations.The main contribution of this paper can be summarized as follows:

\begin{enumerate}
    \item We proposed a new geometric graph neural network architecture GeoGNN for multi-sensory feature extraction and information fusion. By introducing the relative position information into each interaction layer, the model can better incorporate
    obstacle information along with location details from different neighbor robot, and them mapping the fused feature to different moving actions.
    \item Based on GeoGNN, we proposes a new multi-agent collaborative path planning method for real applications. To address the significant disparities between perceptual data and actual sensor data in grid-based approaches, we generate expert data in ROS to train GeoGNN.
    \item We conducted both simulation and physical experiments to validate the effectiveness of the proposed method. The experimental results indicate that GeoGNN achieves good fusion of neighbor information in different expert data and outperforms other traditional graph neural networks in both simulation and real world tests.
\end{enumerate}

\section{Related Work}

\subsection{Graph Neural Networks for Multi-robot problems}

Graph neural networks are commonly employed to extract features from tasks involving multiple robots, which can then be utilized in subsequent algorithms. In \cite{spatical_temporial_learing_TCDS2023}, cascading spatial multi-graph attention was introduced to capture the key spatial relationships among robots. H2GNN \cite{gnn_attention_for_exploration_2022} utilizes a GNN with a multi-head attention mechanism to represent the environment based on observations from different hops of robots and then uses a reinforcement learning algorithm to plan for exploration. Tolstaya et al. use GNN to select points of interest in discretized explored environments \cite{Coverage_sgnn2021}. A completely end-to-end learning framework is proposed in \cite{Visuomotor_rl_2022_liu_TASE2022}, which combines a GNN to represent the environment observed by all robots and a reinforcement learning strategy.

\subsection{Graph Neural Networks for Multi-robot Path Planning}

There are also various graph neural network architectures and methods for processing information from teammates that directly address the MRPP problem using supervised learning. In \cite{mr_collaborativePerception2022}, spatial relationships among robots are encoded into the nodes of GNN via a spatial encoding module. Li et al. represent each robot's observation information, teammates' positions, and the relative location of the destination as a three-channel image. They then use a convolutional neural network to compress the image and utilize GNN to fuse information between different robots further \cite{Li_gcn_decentrral2020, Li_gat_2021}. In the work by authors in \cite{ICRA2023_igbNet}, an uncertainty map channel is added to assess the safety of the neighborhood as input to each node of the graph. They then use their I-GBNet to generate moving actions for the robots.The drawback of directly expressing neighbor position relationships using image channels is the challenge of reconstructing the unique relative position relationship among robots from the channel-based information.



\subsection{Geometric Graph Neural Networks}

A geometric graph is a graph-based data structure in which each node is embedded in an n-dimensional Euclidean space and geometric graph neural networks are a type of architectures that use the spatial relationships contained in geometric graphs to improve their representation ability on geometric graph data, which have achieved great success in fields such as particle physics\cite{ggnn_particlePhysics2020} and quantum chemistry\cite{westermayr2020combining}. 
Various geometric GNNs use different types of relative position information. For example, SchNet\cite{Schnet2018} only uses the relative distance between nodes and ensures coordinate-independent characteristics, while DimeNet\cite{DimeNet2020} uses both distance and relative angle to keep rotation equivariance of the network.

\begin{figure*}
\centering
\includegraphics[width=7in]{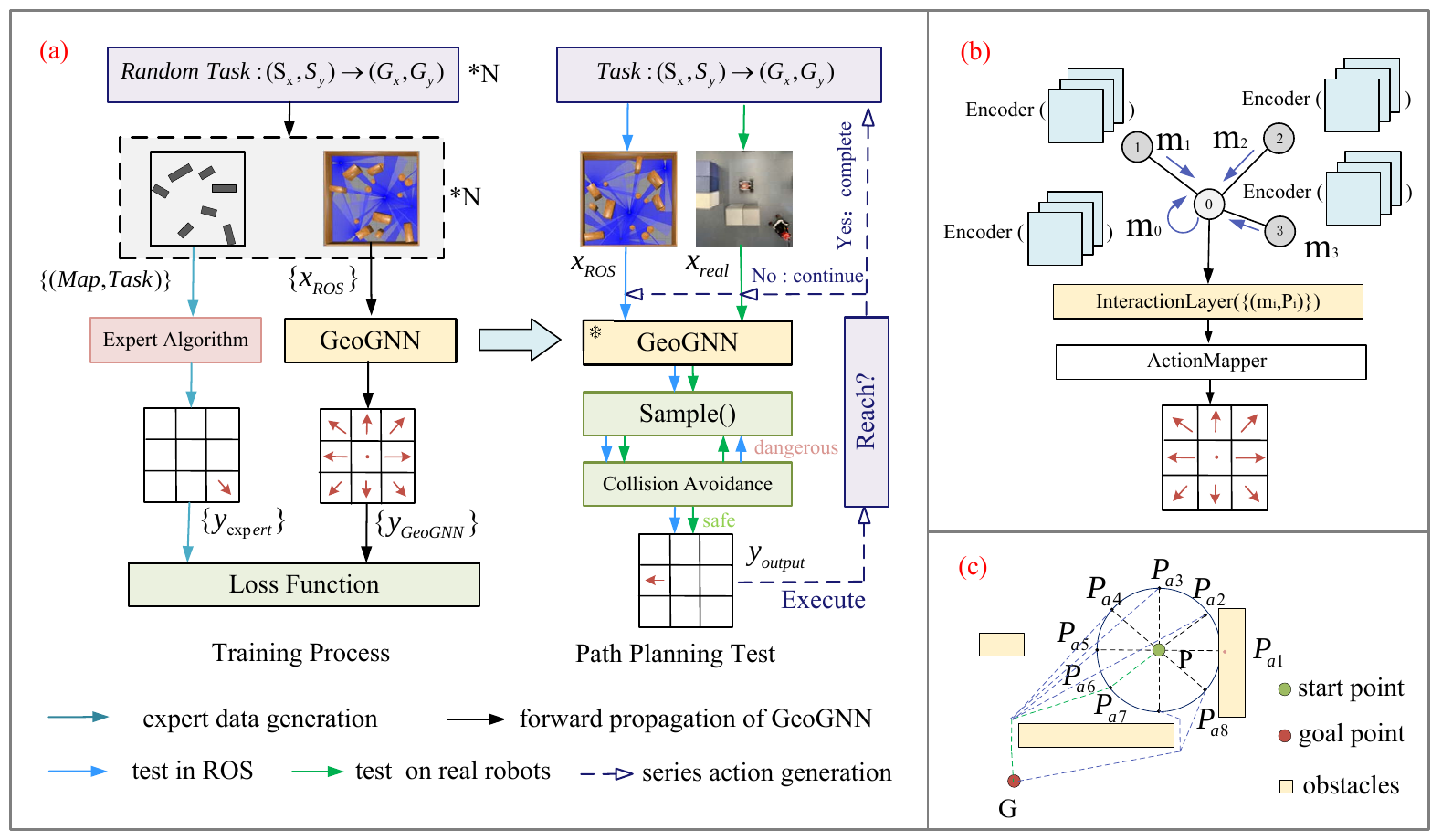}
\caption{Overview of  the  MRPP-GeoGNN framework. (a) Workflow of expert data generation, model training and path planning test in ROS and in the real world. (b) Distributed GeoGNN framework for fusing multi-view sensory data from multi-robot and moving action prediction. (c) Illustration of the expert data generation method.  }
\label{fig:wholeFramework}
\end{figure*}

\section{Problem Formulation}
In this paper, we consider a set of  homogeneous mobile robots $RO=\{R_1,R_2,\ldots,R_N \}$ which reside in 2D environment $E$ with random distributed rectangular obstacles $O=\{o_1,o_2,\ldots, o_M \}$. This study addresses the Point-Goal Navigation problem, where all robots are randomly initialized in obstacle-free area $S=E\setminus O$. The robots need to collectively determine a series of actions to navigate to their individual goal points, which are also randomly selected from $S$.The goal is considered reached when it is within the Field of View (FOV) of the corresponding robot; when out of view, the robot can determine its direction.

At any time in their episode, each robot can take an action from nine standard actions $A=\{(cos(n\pi/4),sin(n\pi/4)| n\in \{0,1,\ldots,7\}\} \cup\{(0,0)\}$ that move itself towards one of eight directions or stop where it is.
Furthermore, each robot is unable to access to global position and only has local Field of View(FOV) with a fixed radius $r_{FOV}$, within which it is able to detect obstacles and teammates,  and it can communicate with teammates whose distance are within effective communication range $r_{COM}$.

Under these assumptions, the sensory information of all robots and the communication connection status among robots can be formulated as a geometric graph$\quad \mathcal{G} = (\mathcal{V},\mathcal{E})$ with $\mathcal{V}=\{v_1,v_2,\ldots v_n \},v_i\in R^F$ denotes the local observations of all robots and $e_{i,j}=(r_{ij},\theta_{ij}))\in \mathcal{E}$ if $r_{ij}<r_{COM}$, 
where $r_{ij}$ is the relative distance between robot $i$ and robot $j$ , $\theta_{ij}$ is  angle of robot $j$ relative to robot $i$.When $r<r_{COM}$, robots can communicate with each other, and there is a corresponding edge between two nodes in $\mathcal{G}$.

\section{Methodology}

In this study, we introduce MRPP-GeoGNN, a novel collaborative path planning approach for multiple mobile robots in real-world applications, as illustrated in Figure \ref{fig:wholeFramework}. Within this framework, each robot utilizes GeoGNN to merge local information, making decisions on movement directions or pausing until reaching its designated goal point after a sequence of movement actions. Each movement action entails the robot relocating a fixed distance.The neural network is trained in a supervised manner to learn knowledge from expert datasets regarding the optimal choice of movement action based on the local environment perceived by the target robot and its neighboring peers. To bridge the disparity between Lidar data and simplified grid matrices that disregard factors like Lidar occlusion by obstacles, we leverage the Robot Operating System (ROS) to simulate Lidar data for model training and devise a methodology for generating expert datasets within ROS to enable the trained model's utilization in both ROS simulations and actual robotic platforms.

\subsection{Network architecture}
MRPP-GeoGNN is a distributed execution architecture which utilizes the spatial relationship between robots in the higher-level feature abstraction process to achieve better collaboration among robots.We denote the observations from sensors of robots as $\{x_i\}$ and the output of MRPP-GeoGNN as $\{y_i\}$ respectively, where $i \in IDs=\{1,2,\ldots,N\}$ and $y_i \in A $. Let $\mathcal{N}(i)$ denotes robots that can communicate with robot $i$. The actual execution process of GeoGNN for path planning is shown in figure \ref{fig:wholeFramework}(b). After acquiring sensory data, the encoder compresses it into observation features which are transmitted to neighboring robots through a communication device. This approach significantly alleviates the communication burden between robots compared to transmitting raw data, eliminating the need for redundant calculations across multiple robots.The core of GeoGNN,  which is composed of interaction layers that are connected sequentially, then fuses location-related messages received from neighbors with the feature compressed by encoder of the robot itself. In GeoGNN, relative position data is used in each of the interaction layer to extract spatially related features. Eventually, the ActionMapper maps the feature extracted by GeoGNN into actions that can be taken by robots.

\subsubsection{\textbf{Sensory data encoder}}
We transform the sensory data of each robot into a three-channel map, as described in \cite{LiQb_2020GNN}: the local map channel, neighbor position channel, and goal position channel. The local map channel utilizes distance data from radar to construct a $d\times d$ matrix, where the boundaries of the field of view are represented by 1 and other areas by 0. The neighbor channel and goal channel designate the positions of teammates within the view of the target robot and the position of its goal using the number "1". If the goal point is outside the field of view, the boundary points of the goal matrix are set to 1 to indicate the direction in which the robot should move.

Convolutional neural network(CNN) is used to extract useful features $f\in R^F$ from aforementioned three channels local map and then transfer these messages to neighbor robots.
In this work, a mini VGG\cite{simonyan2014VGG} is used to extract features from three-channels map , which is composed of sequentially blocked  Conv2d-BatchNorm2d-ReLU-MaxPool2d and Conv2d-BatchNorm2d-ReLU for three times, finally a fully connected layer is used to project the flattened CNN feature into a F-dimensional vector.

\subsubsection{\textbf{Interaction layer}}

In GeoGNN, we use several sequentially connected interaction layers to extract fusion features from local sensory information at different abstraction levels. Denote the number of interaction layers as $h$, then GeoGNN with h interaction layers can  fuse the sensory information of all robots within the h-hop range neighbor nodes on graph $\mathcal{G}$.The embedding in each interaction layer is updated using messages from neighbors combined with the relative location of robots shown in equation (\ref{eqn:message_fusion}).
\begin{equation}
\begin{aligned}
    &v_i^l = AGG(f_i^l,\{(f_j^l,p_{ij}^l)| j\in \mathcal{N}(i) \}) \\
    &f_i^{l+1}=UPD(f_i^l,v_i^l)	    
\end{aligned}
    \label{eqn:message_fusion}
\end{equation}

\begin{figure}[t]
\centering
\includegraphics[width=1.8in]{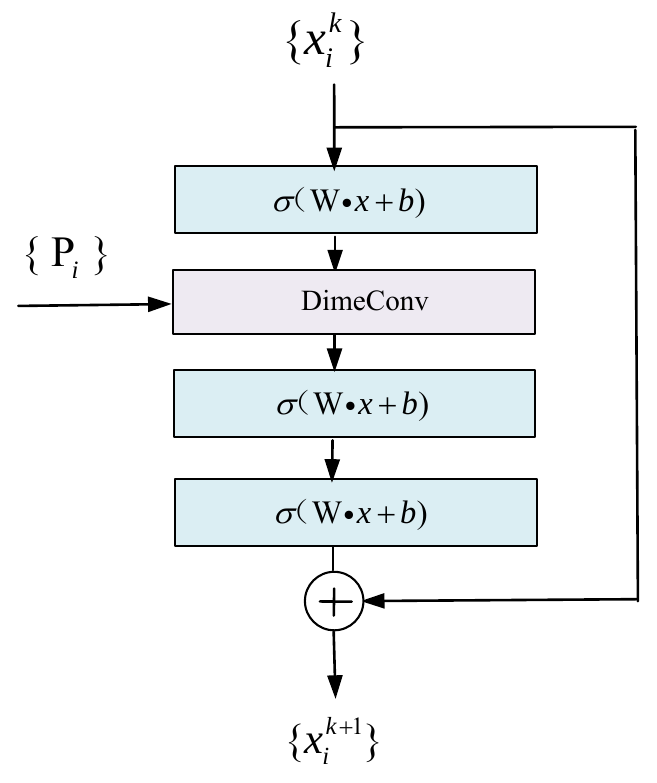}
\caption{Interaction layer}
\label{fig:interactionLayer}
\end{figure}

\begin{figure}[t]
\centering
\includegraphics[width=3.0in]{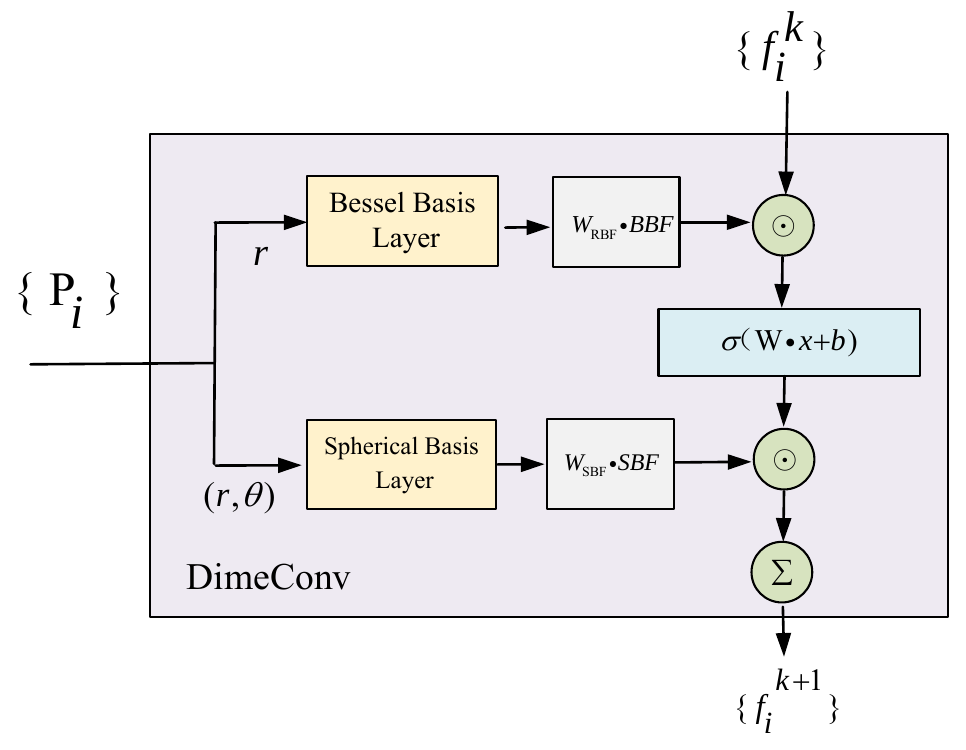}
\caption{DimeConv layer}
\label{fig:DimeConvLayer}
\end{figure}

The structure of interaction layer is shown in figure \ref{fig:interactionLayer}. The aggregation function is realized by a series of linear layer, Shifted Soft-plus Layer, and a DimeConv Block that encodes relative position information into spatially related feature. 
The detail realization of DimeConv Block is shown in  \ref{fig:DimeConvLayer}. 
The update function $UPD(\cdot)$ is equipped with a ResNet-style in equation \ref{eqn:message_update}, which enables features extracted by lower layers to be further used at high levels of feature extraction.Shifted Softplus layer has the nonlinear function $ssp(x)=\log (0.5e^x+0.5)$.
\begin{equation}
    f_i^{l+1}=f_i^l+v_i^l
    \label{eqn:message_update}
\end{equation}

\subsubsection{\textbf{DimeConv Module}}


Unlike SchNet\cite{Schnet2018} that only uses the relative distance between two nodes in the radial basis function,  we use Bessel Basis function and Spherical Basis function \cite{DimeNet2020} to acquire the spatial feature between two robots, this is because in the context of multi-robots path planning, both relative distance and relative angle of neighbor robots are crucial to making decisions given their sensory information. 
The Bessel Basis function(BBF) uses the relative distance shown in Equation \ref{eqn:SBFnBBF}, where $c$ is the cutoff distance and we set $c=r_{COM}$. In Spherical Basis function(SBF) , $z_{ln}$ is the $n^{th}$ root of $l$-order Bessel function, $j_l(x)$ and $Y(x)$ represent the first kind spherical Bessel functions and spherical harmonics respectively.
\begin{equation}
    \begin{aligned}
        &BBF_n(r)=\sqrt{\frac{2}{c}}\frac{sin(\frac{n\pi}{c}r)}{r}, n=1,2,\ldots,N_{RBF} \\
 & SBF_{nl}(r,\theta)=\sqrt{\frac{2}{c^3} j_{l+1}^2(z_{ln})j_l(\frac{z_ln}{c}r) Y_l^0 (\theta)}
    \end{aligned}
    \label{eqn:SBFnBBF}
\end{equation}

After getting the SBF and SBF value of neighbors, we use two dense layers without bias to map their dimensions to be the same as the neighbor feature dimensions and then perform Hadamard product operation (marked as $\odot$) successively, the Linear-ShiftedSoftPlus modules are added to reduce coupling between different dimensions and enhance the expressive ability of models. 

\subsubsection{\textbf{Action Mapper}}

After update of internal feature with GeoGNN using neighbor features and relative locations, a multilayer perceptron network is used to output the probability of all actions according to fused feature. It is composed of two Linear layers and a Nonlinear layer. Given the output of the $MLP:o_1,o_2,\ldots ,o_n,$  the action generated by the whole neural network is calculated by the softmax function as Equation \ref{eqn:output}.
\begin{equation}
    y = argmax(softmax({o_i})) = argmax(o_i)
    \label{eqn:output}
\end{equation}

\subsection{Expert data generation method}
In order to better simulate the movement of robots in the real world, we designed an expert data generation scheme. The estimated path length in each direction is the sum of the following two parts: (1) fixed forward step length $l_{step}$ (2) expert path length from the arrival point to the target point after moving forward at a fixed step length. When a fixed step forward collides with an obstacle, the estimated path length in this direction is infinite. The direction with the minimum estimated length in all forward directions will be used as the expert action label $y$ of the robot at that position. As shown in figure \ref{fig:wholeFramework}(c), $G$ is the destination of the robot, $ P $is the current position of the robot, and $P_{ai}$ is the position reached by the robot after advancing a fixed step $l_{step}$ in the i-th direction, so the estimated path length in the i-th direction can be computed according to Equation \ref{Eqn: estmated_length}. And the action label $y$ given by the expert algorithm to position $P$ is given by Equantion \ref{Eqn:expert_label}.
 \begin{equation}
   \hat{l_i} = l_{step} + expert(P_{a_{i}}, G)
   \label{Eqn: estmated_length}
 \end{equation}

\begin{equation}
    y = argmin_{i}\{\hat{l_i}\}
    \label{Eqn:expert_label}
\end{equation}

\subsection{Learning from expert data}
We train our models in a supervised way and we use $A^*$ algorithm to generate action labels for each robot. During the expert data generation process, the presence of the global map matrix ensures that the actions taken by the expert are globally optimal. The expert dataset is divided into training, testing, and validation sets in a ratio of 3:1:1.
In the training process, we have access to the globally optimal action for each robot performing a Point-Goal task.  The training objective is to acquire a mapping $M(\cdot)$ so that the output is as close as possible to the corresponding action labels $Y$ from expert data with the input $X$ being the sensory data of all robots. We use a cross entropy loss $\mathcal{L}(\cdot)$ as the objective function with trainable parameters $\theta$ of our model:
 \begin{equation}
 \hat{\theta} = argmin_{\theta} \mathcal{L}_{\theta}(M(X),Y)
 \end{equation}

 \subsection{Sequential action generation}
 In the actual execution process, the robot keeps communicating with its neighbors and transmits the features extracted from its own perception information and the feature output of all K-1 interaction layers, and finally obtains the recommended actions from GeoGNN. However, the robot cannot directly adopt this action and needs to go through the following three steps to obtain the final safe action: (1) Static obstacle avoidance: When the robot's action will collide with a static obstacle, the robot will resample from the action space. (2) Dynamic obstacle avoidance: When the action taken by the robot may collide with a higher-level neighbor, the action needs to be resampled. In this article, the priority of a robot is determined simply by its index. (3) Avoidance of local loops: When the action taken by the robot will cause it to perform reciprocating motion in a certain local area, the robot needs to reselect its action. The selected action after the above processing will be output as the action actually performed by the robot.

    

\section{Experiments}
We setup a series of simulation experiments and physical tests to evaluate the effectiveness of the proposed method. 

\subsection{Experimental setup}

\subsubsection{\textbf{Expert dataset}}
\begin{figure}[!t]
\centering
\includegraphics[width=3.4in]{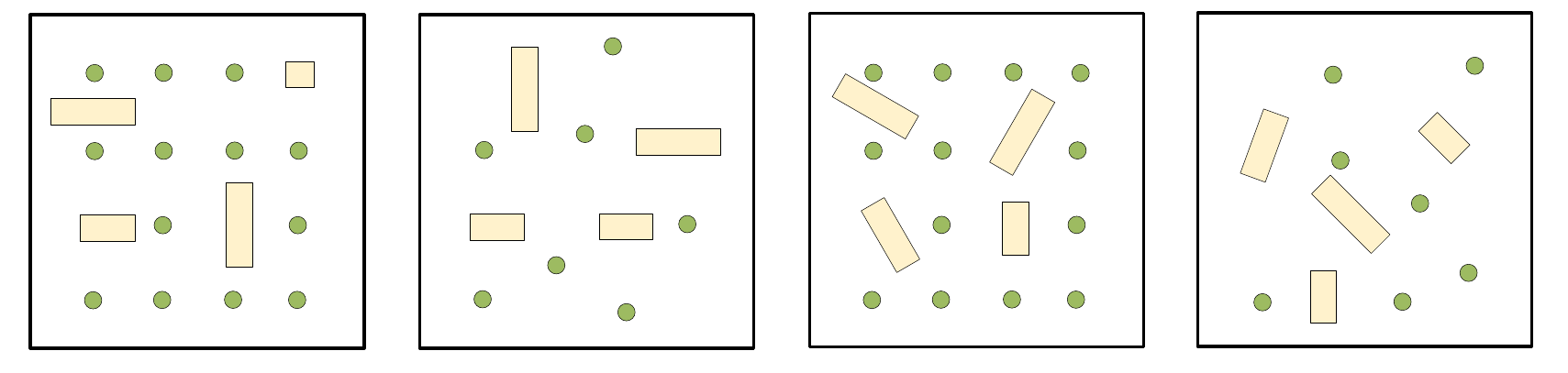}
\caption{Four different types of expert datasets generated.}
\label{fig:DataSets}
\end{figure}

In order to better simulate the real environment, we use ROS to generate different types of expert data. As shown in figure \ref{fig:DataSets}, two different types  of maps (simple map and complex map) and two different robot distribution types (fixed interval distribution or random distribution) are combined, forming four types of expert data: $\mathcal{D}_{simple-grid}$ ,$\mathcal{D}_{complex-grid}$, $\mathcal{D}_{simple-random}$ , $\mathcal{D}_{complex-random}$. In simple maps, the angle of any rectangular obstacle $\theta_{obs}$ is chosen from binomial distribution $\theta_{obs} \sim \pi/2 \times B(2,0.5)$, while in  complex maps,  $\theta_{obs}$ is chosen from uniform distribution :  $\theta_{obs} \sim U(0,2\pi)$. Robots are placed randomly in $\mathcal{D}_{simple-random}$ and $\mathcal{D}_{complex-random}$, while  in  $\mathcal{D}_{simple-grid}$ and $\mathcal{D}_{complex-grid}$ , robots are placed in obstacle-free areas at a fixed distance $d$ and robots will have 8 neighbors when $\sqrt{2}d < r_{COM} < 2d $ and 4 neighbors when $ d < r_{COM} < \sqrt{2}d$. All of these datasets utilize expert algorithms to calculate optimal paths for robots from their initial positions to randomly selected goal points in obstacle-free areas.

The map size is $20\times20$ $m$, and its resolution is set to $0.05m$ when represented as a matrix. Obstacles in the map are randomly chosen from three types: $1/2/3m \times 0.5m$ and randomly placed with orientations chosen from different distributions.The maximum detectable distance of the radar $r_{FOV}$ is $5m$, whose angular resolution is $1$\textdegree thus raw data from radar are $360$-dimensional vectors. The number of robots $N$ is $15$ in $\mathcal{D}_{random}$ and the interval between two robots in $\mathcal{D}_{grid}$  and $\mathcal{D}_{row}$ is $3m$.

\subsubsection{\textbf{Model training and testing details}}
In the training process, Adam optimizer is used with a learning rate of $10^{-3}$, a weight decay of $10^{-4}$ and the batch size of each iteration is 16, ReduceLROnPlateau scheduler is used to reduce learning rate when the loss has stopped lowering. We train 1000 iterations in each epoch on the train set and 100 epochs for the whole training process. Each model was trained for 5 times to get the maximum accuracy on the test dataset.

\begin{figure}[!t]
	\centering  
	\subfigcapskip=-5pt 
        \subfigure[case 1]{
		\includegraphics[width=0.48\linewidth]{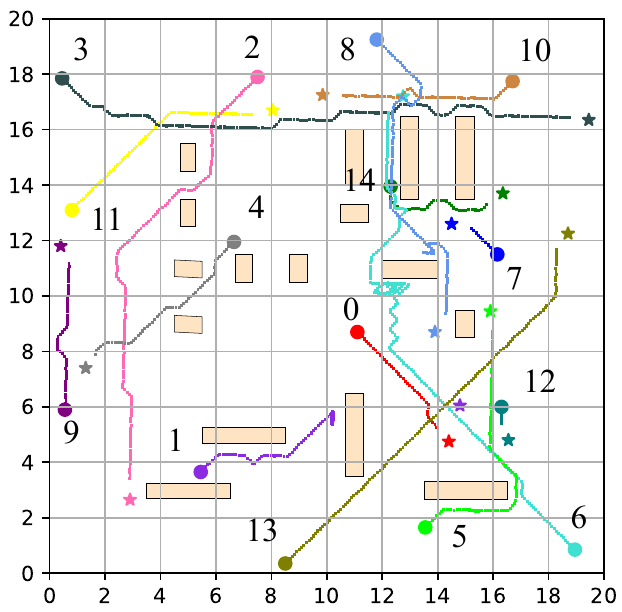}} 
	\subfigure[case 2]{
		\includegraphics[width=0.48\linewidth]{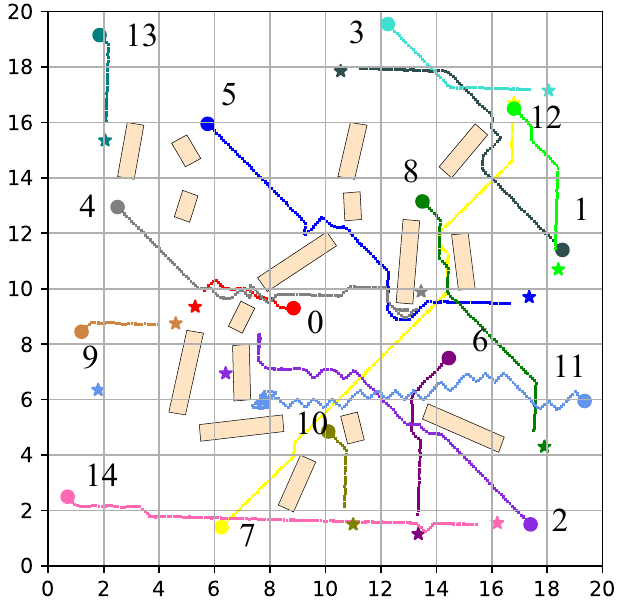}}\\
          \subfigure[case 3]{
		\includegraphics[width=0.48\linewidth]{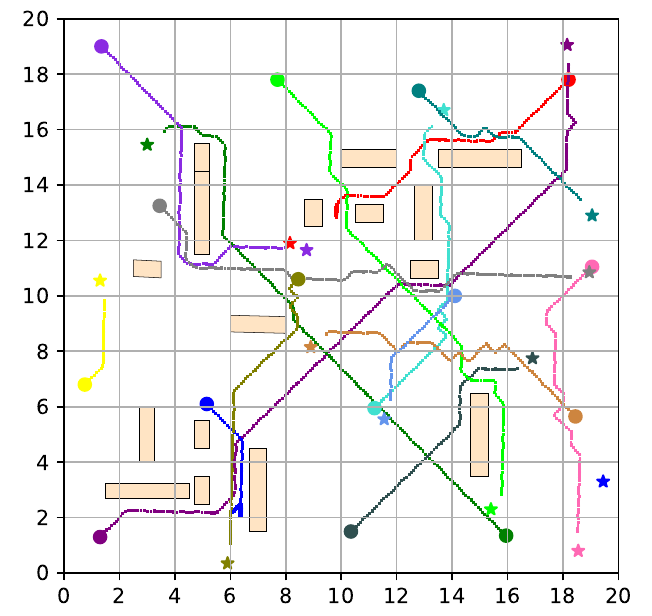}} 
          \subfigure[case 4]{
		\includegraphics[width=0.48\linewidth]{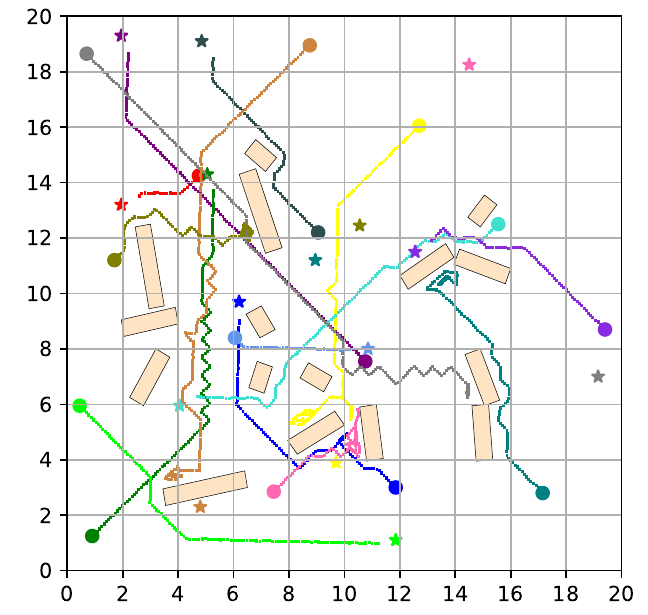}} 
        
	\caption{Four test cases in ROS.}
        \label{fig:tsne}
\end{figure} 

\begin{figure}[!t]
\centering
\includegraphics[width=2in]{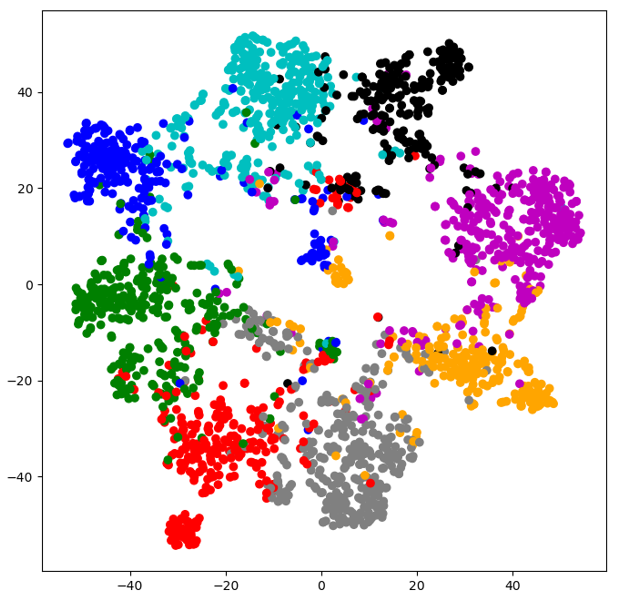}
\caption{TSNE analysis result on test set. GeoGNN is trained on $\mathcal{D}_{complex-random}$, and hop=2.}
\label{fig:tsne}
\end{figure}

\begin{table*}[hb]
\renewcommand\arraystretch{1.5}
\centering
\caption{Accuracy($\uparrow$ of different models on four expert datasets. $r_{COM}=5$ }
\label{table:acc}
\begin{tabular}{ccccccccccccc}
\toprule 
\multirow{2}{*}{Model} & \multicolumn{3}{c}{$\mathcal{D}_{simple-grid}$} & \multicolumn{3}{c}{$\mathcal{D}_{simple-random}$} & \multicolumn{3}{c}{$\mathcal{D}_{complex-grid}$} & \multicolumn{3}{c}{$\mathcal{D}_{complex-random}$}  \\
\cline{2-13}
                & $1hop$ & $2hops$ & $3hops$ & $1hop$ & $2hops$ & $3hops$
                & $1hop$ & $2hops$ & $3hops$ & $1hop$ & $2hops$ & $3hops$
                \\
\midrule
CNN             & 0.786 & -     & -     & 0.780 & - & - 
                & 0.775 & -     & -     & 0.764 & - & - \\
GraphSAGE       & 0.835 & 0.830 & 0.818 & 0.810 & 0.805 & 0.807
                & 0.802 & 0.805 & 0.798 & 0.791 & 0.784 & 0.788 \\
MAGAT           & 0.821 & 0.824 & 0.819 & 0.799 & 0.804 & 0.806 
                & 0.783 & 0.776 & 0.789 & 0.781 & 0.769 & 0.784\\
g-GNN           & 0.827 & 0.831 & 0.833 & 0.808 & 0.813 & 0.803 
                & 0.799 & 0.784 & 0.805 & 0.790 & 0.800 & 0.792\\
       
GeoGNN          & 0.840 & 0.847 & 0.850 & 0.819 & 0.827 & 0.825 
                & 0.810 & 0.811 & 0.817 & 0.793 & 0.806 & 0.805 \\
\bottomrule
\end{tabular}
\end{table*}

\begin{table*}[hb]
\renewcommand\arraystretch{1.5}
\centering
\caption{Flowtime Increase($\%$) ($\downarrow$) and Success Rate ($\uparrow$) of test results on 500 random maps in ROS. }
\label{table:flowtime}
\begin{tabular}{cccc|ccc|ccc|ccc}
\toprule 
\multirow{3}{*}{Model} & \multicolumn{6}{c}{Simple Map} & \multicolumn{6}{c}{Complex Map}  \\
\cline{2-13} & \multicolumn{3}{c}{FI} & \multicolumn{3}{c}{SR} & \multicolumn{3}{c}{FI} & \multicolumn{3}{c}{SR} \\ 

\cline{2-13}
                & $1hop$ & $2hops$ & $3hops$ & $1hop$ & $2hops$ & $3hops$
                 & $1hop$ & $2hops$ & $3hops$ & $1hop$ & $2hops$ & $3hops$\\
\midrule
CNN             & $42.9 $ & -     & -     & 0.656 & -     & -    
                & $47.4 $ & - & -   & 0.631 & - & -  \\
GraphSAGE       & $34.9$ & $36.1 $ & $36.5$ & 0.692 & 0.684 & 0.680 
                & $40.2 $ & $42.1  $ & $44.7 $ & 0.660 & 0.648 & 0.651 \\
MAGAT           & $36.1  $ & $38.4 $ & $37.3  $ & 0.653 & 0.662 & 0.687
                & $43.1  $ & $44.5 $ & $45.8 $  & 0.645 & 0.664 & 0.656\\
g-GNN           & \textbf{34.2}& $35.9 $ & $37.2$ & 0.689 & 0.697 & 0.690
                & \textbf{38.6}  & $42.2  $ & $39.9 $  & 0.658 & 0.673 & 0.669 \\
    
GeoGNN          & $34.4$ & \textbf{32.7} & \textbf{33.0} & \textbf{0.700} & \textbf{0.705} & \textbf{0.719}
                & $39.7$ & \textbf{39.2}  & \textbf{38.8}  & \textbf{0.688} & \textbf{0.679} & \textbf{0.680}\\
\bottomrule
\end{tabular}
\end{table*}

 When testing in ROS, we generate 500 random maps and for each map, we randomly generated initial positions for all robots and set their own random goals.The total path lengths $l$ of all robots during their lifetime $t$ are recorded.

In the physical experiments, five mobile robots were used, and an off-site UWB system was used to establish a global coordinate system, which was then transformed into a local coordinate system for each individual robot during operation. The YOLOv3\cite{2018YOLOv3} algorithm was used to identify teammates within the field of view. Each robot was equipped with a laser radar and an  integrated graphics card to support the computation. The extracted features were transmitted to teammates within the field of view via UWB. We conducted 10 sets of physical experiments, with each robot being assigned random start and end points in a 7x7 meter experimental field.

\subsubsection{\textbf{Baselines}}
Regarding performance on the expert dataset, we introduce three GNN-based models and one CNN-based metods as our baselines, all of these baselines have the same sensory data encoder and maintain internal features in the CNN or GNN layers with a dimension of 128.

\begin{itemize}
    \item[1)] \textit{GraphSAGE}: The main difference of this framework is that it utilizes GraphSAGE \cite{NIPS2017_GraphSAGE} to process messages from teammates.
    \item[2)] \textit{MAGAT} \cite{LiQb_2021GAT}: The framework, named MAGAT, uses a key-query-like attention mechanism to select important messages from neighbors.
    \item[3)] \textit{g-GNN}\cite{mr_collaborativePerception2022}: A GNN-based framework from multi-robot collaborative perception network, here we use the network in \cite{mr_collaborativePerception2022}  to replace the message fusion module to get action output.
    \item[4)] \textit{CNN}: We use only the Mini-VGG for comparison, which has no interaction layers that enable robots to communicate with neighbors, using only the observation information of a single robot itself to generate actions.
\end{itemize}

\subsubsection{\textbf{Evaluation Metrics}}
Following metrics are used to evaluate the performance of algorithms:

\begin{itemize}
    \item[1)]  Accuracy: \textit{$Acc = n_{right}/n$}. Given that the goals of all robots and obstacles are randomly chosen from obstacle-free area, we use the accuracy of models on all classes to evaluate the performance of algorithms on expert data. $n$ is the total number of predictions and $n_{right}$ is the total number of correct predictions across all categories of action labels.
    \item[2)]  Flowtime Increase: $FT=avg((l-l_{expert})/l_{expert})$, expresses the percentage change between the returned path length $l$
    and the expert’s execution path length. $avg(\cdot)$ is the average function exerted on all robots across all maps tested.
    \item[3)] Success Rate: $SR=n_{reach}/n$,  used to measure the ability of different algorithms to successfully plan paths within a given period of time, where $n_{reach}$ is the number of robots that reach their destination durting their lifetime and $n$ is total number of robots that have been test in all maps.
\end{itemize}

\subsection{Model performance on expert datasets}
To evaluate the model's ability to imitate expert algorithms, we trained and tested the model on four types of expert data.
As shown in Table \ref{table:acc}, GeoGNN achieves better performance across expert datasets compared to baselines.

When robots are distributed at fixed intervals in the map, their neighbor robots tend to appear at specific relative positions. Therefore, GNN-based models are able to capture this characteristic and achieve better results compared with random distribution.Specifically, the ability of GeoGNN to learn this distribution characteristic is more prominent than other GNN-based baselines, whose accuracy is about $5\%$ higher than that of CNN in grid-like data.
When the obstacles in the map are oriented more randomly, the actual environment becomes more complex, and it becomes more difficult for the models to learn the correct actions. In this condition, GeoGNN can still achieve about $80\%$ accuracy on complex maps with random distributed robots.

Table \ref{table:acc} also shows the performance differences between graph neural network models with different numbers of interaction layers.When the number of interaction layers is more, the number of hops to fuse neighbor information is also more, and theoretically the model can achieve better performance.But we failed to see this phenomenon in some baselines, and the performance of some models even deteriorated. This may be because when the number of robots is limited, the probability that distant neighbors appear at key locations is smaller, making it more difficult to extract effective information from these neighbor information.

\subsection{ROS test results}
To evaluate the actual performance of the models, we tested them on aforementioned two different types of maps in ROS. 500 maps of each type are randomly generated, and the number of robots in each map is 15. All robots are randomly born, and their goals are randomly chosen from obstacle-free area in maps. The statistical results of flowtime increase and success rate are presented in Table \ref{table:flowtime}, which confirms the effectiveness of our proposed algorithm.

As shown in Table \ref{table:flowtime} ,compared with other baseline algorithms, our proposed algorithm has the lowest flowtime value relative to the path length of the expert algorithm.In simple maps, when the hop of interaction layer is 2, GeoGNN has the lowest flowtime increase value $32.7\%$ and the highest reach rate $71.9\%$ when $hop=3$.And in complex maps, when the communication hop number is 3, GeoGNN reaches the lowest flowtime increase value $38.8\% $ and the highest reach rate $68.8\%$ when  $hop=1$.

\subsection{Ablation study}
In order to verify the effectiveness of the fusion method of relative position information used in the interaction layer, we conducted ablation experiments, and the results are shown in Table \ref{table:Ablation_Study}. We compared three models: GeoGNN that does not add a neighbor relative position information module to interaction layers(GeoGNN w.o. Nghb), GeoGNN that uses radial basis functions(RBF) to process relative distances and then integrates the information into interaction layers(GeoGNN w. RBF), and our model (GeoGNN w. BBF-SBF). Experimental results show that the performance of models that do not incorporate neighbor relative position information or only incorporate neighbor relative distance information in the interaction layer is significantly weakened, which confirms the effectiveness of the proposed method.

\begin{table}[!t]
\renewcommand\arraystretch{1.5}
\centering
\caption{Abalation Study of different modules in the interaction layer, hop=2}
\label{table:Ablation_Study}
\begin{tabular}{ccccccc}
\toprule 
\multirow{2}{*}{GeoGNN} & \multicolumn{3}{c}{Simple Map} & \multicolumn{3}{c}{Complex Map} \\
\cline{2-7}
                &{\scriptsize Acc $ \uparrow $} & {\scriptsize $FT \downarrow$} & {\scriptsize SR$\uparrow $} 
                &{\scriptsize  $ Acc\uparrow $} & {\scriptsize $FT \downarrow$} & {\scriptsize SR$\uparrow $}
                \\
\midrule

 w.o. Nghb       & 0.800 & 35.6$\%$ & 0.678 & 0.786 & 42.4$\%$   & 0.660 \\
 w. RBF          & 0.811 & 34.5$\%$ & 0.688 & 0.794 & 41.6$\%$ & 0.654\\
 w. BBF-SBF      & 0.827 & 32.7$\%$ & 0.705 & 0.806 & 39.2$\%$ & 0.679 \\
\bottomrule
\end{tabular}
\end{table}

\begin{figure}[!t]
\centering
\includegraphics[width=3.4in]{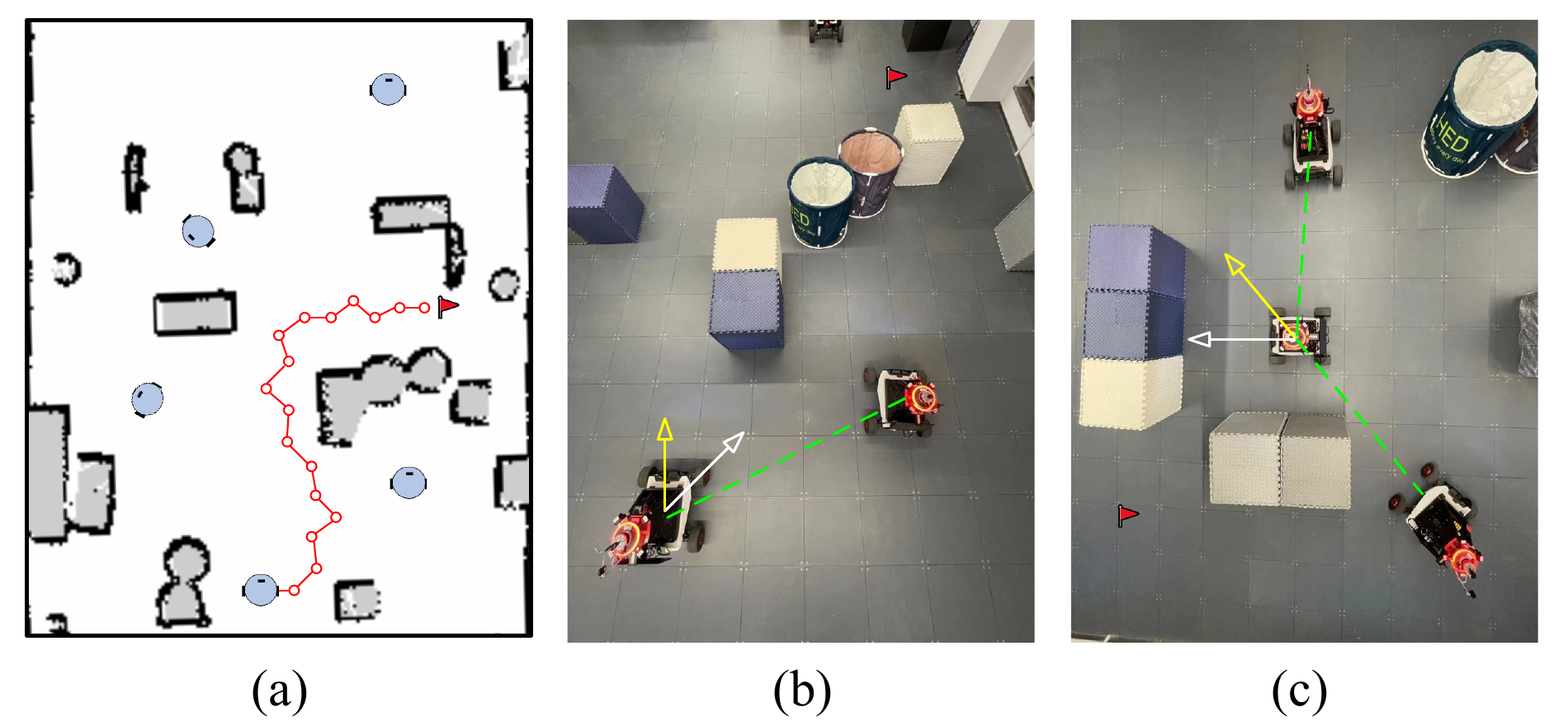}
\caption{Physical experimental results. (a) shows the path of the target robot in a case, (b) and (c) shows the comparison of the decision results of CNN and GeoGNN in different scenarios.(white arrow : CNN output,  yellow arrow :  GeoGNN output,  green dotted line : teammate communication)}
\label{fig:physicalCase}
\end{figure}

\begin{figure}[!t]
\centering
\includegraphics[width=3in]{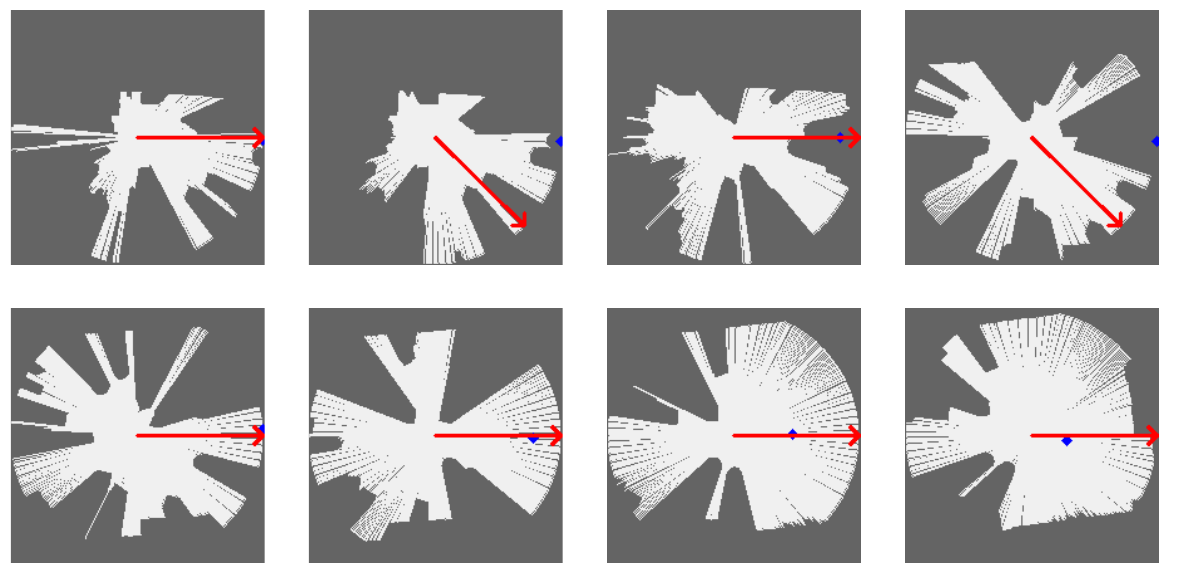}
\caption{The important decision-making points of the target robot in the process of moving in the physical experiment, where the red arrow is the forward direction of the GeoGNN output, and the blue dot is the target direction.}
\label{fig:actionSeries}
\end{figure}

\subsection{Physical Experiment}
The experimental results showed that the average path length based on the GeoGNN method was 8.79 meters, while the average path length based on CNN was 9.25 meters. The results confirmed the effectiveness of the proposed method in integrating neighbor information. Figure 7 shows the motion trajectory of a robot in the experiment, with (b) and (c) showcasing two effective experimental scenarios where GeoGNN utilized neighbor information, making better decisions. In scenario 7a, the target direction was obstructed by obstacles, causing the CNN-based direction to head towards the side with obstacles, making it difficult to avoid them. In contrast, the GeoGNN-based method effectively utilized the perception information of the teammate in this area and selected a more efficient forward direction at a further distance.

\section{Conclusion}
In this paper, we propose a method for multi-agent collaborative path planning based on geometric graph neural networks, which is implemented by incorporating spatial relationship  of neighbor robots into the interaction layer to achieve better results. It can effectively integrate the sensing information of neighbors at different locations within different communication hop ranges, thereby better simulating expert algorithms for path planning.An expert data generation method for robot single-step movement in ROS environment is proposed, by which we generate expert data for training and testing of GeoGNN. Experimental results show that the accuracy of the proposed method is improved by about $5\%$ compared to the model based only on CNN on the expert datasets. In the ROS simulation environment path planning test, the success rate is improved by about $4\%$ compared to CNN-based model and the flowtime increase value is reduced by about $18\%$, which is better than other baseline graph neural network models. In the future, we will further integrate temporal relationships and improve existing methods to further enhance the model's actual path planning capabilities.

\ifCLASSOPTIONcaptionsoff
  \newpage
\fi

\bibliographystyle{IEEEtran}

\bibliography{ref}

\end{document}